\documentclass{article}

\usepackage[final]{neurips_2025}

\usepackage[utf8]{inputenc} 
\usepackage[T1]{fontenc}    
\usepackage{hyperref}       
\usepackage{url}            
\usepackage{booktabs}       
\usepackage{amsfonts}       
\usepackage{nicefrac}       

\usepackage{microtype}      
\usepackage{xcolor}         
\usepackage{enumitem}
\usepackage{xcolor}
\usepackage{amsmath}
\usepackage{cleveref}
\usepackage{algorithm}
\usepackage{algpseudocode}
\usepackage{amssymb}
\usepackage{wrapfig}
\usepackage{tcolorbox}
\usepackage{amsthm}
\usepackage{booktabs} 
\usepackage{xcolor}  

\definecolor{mildred}{RGB}{200, 60, 60}
\definecolor{mildblue}{RGB}{70, 130, 180}   
\definecolor{mildgreen}{RGB}{85, 170, 85}

\title{Algebraformer:  A Neural Approach to Linear Systems}

\author{%
  Pietro Sittoni \\
  Gran Sasso Science Institute \\
  \texttt{pietro.sittoni@gssi.it} \\
  \And
  Francesco Tudisco \\
  University of Edinburgh \& Miniml.AI \\
  \texttt{f.tudisco@ed.ac.uk} \\
}

\begin{document}

\maketitle

\begin{abstract}
Recent work in deep learning has opened new possibilities for solving classical algorithmic tasks using end-to-end learned models. In this work, we investigate the fundamental task of solving linear systems, particularly those that are ill-conditioned. Existing numerical methods for ill-conditioned systems often require careful parameter tuning, preconditioning, or domain-specific expertise to ensure accuracy and stability. In this work, we propose Algebraformer, a Transformer-based architecture that learns to solve linear systems end-to-end, even in the presence of severe ill-conditioning. Our model leverages a novel encoding scheme that enables efficient representation of matrix and vector inputs, with a memory complexity of $\mathcal{O}(n^2)$, supporting scalable inference.
We demonstrate its effectiveness on application-driven linear problems, including interpolation tasks from spectral methods for boundary value problems and acceleration of the Newton method. Algebraformer achieves competitive accuracy with significantly lower computational overhead at test time, demonstrating that general-purpose neural architectures can effectively reduce complexity in traditional scientific computing pipelines.
\end{abstract}

\section{Introduction}

Linear systems are ubiquitous in the physical sciences. Many fundamental tasks in physics, from solving inverse problems \cite{Groetsch2011} to performing interpolation \cite{lunardi2018interpolation}, or from discretizing and simulating differential equations \cite{butcher1964implicit, trefethen2000spectral}, ultimately reduce to solving a system of linear equations. Yet, while the theory behind linear solvers is well established, the practical numerical solution of \emph{ill-conditioned} systems remains notoriously challenging. These systems are extremely sensitive to input perturbations and often lead to numerical instability, making their resolution highly dependent on problem-specific choices such as solver type, preconditioner design, and regularization strategy. As a result, even standard scientific computing libraries may yield poor results unless the system structure is explicitly accounted for and, in general, approaching these problems requires substantial expertise in numerical linear algebra, creating a barrier to accessible and robust solutions.

Over the decades, three main classes of methods have been developed. \emph{Direct methods} (LU, Cholesky, QR \cite{davis2016survey}) are accurate but computationally prohibitive for large or sparse systems. \emph{Iterative solvers} (Krylov methods \cite{van2003iterative}, Jacobi, Gauss-Seidel, SOR \cite{saad2003iterative}) scale better but require tuning and often fail under ill-conditioning. \emph{Randomized algorithms} \cite{martinsson2020randomized,strohmer2009randomized} offer scalability but remain sensitive to conditioning. In all cases, the condition number is central, motivating preconditioning \cite{chen2005matrix,meijerink1977iterative,benzi2002preconditioning,brandt1977multi} and regularization \cite{tikhonov1963solution,Phillips_ref,davoli2019adaptive,saksman2009discretization,Gilboa_Guy,rudin1992nonlinear,Total_Generalized_Variatio,shumaylovweakly}, which improve stability but demand careful design.

In this work, we propose a new machine learning-based paradigm for solving linear systems that is simple, general, and effective even in the presence of severe ill-conditioning. Inspired by recent advances in machine learning, end-to-end solution for scientific computing, from solving PDEs \cite{azizzadenesheli2024neural, brandstetter2022message} to learning classical algorithms \cite{pmlr-v162-velickovic22a, velikovi2019neural, DBLP/corr/abs-2409-07154}, we introduce \textbf{Algebraformer{}}, a Transformer-based architecture \cite{vaswani2017attention} that can solve linear systems via a single forward pass. 

Overall, \textbf{our main contributions are as follows}:
\begin{itemize}[leftmargin=*,noitemsep=0pt,topsep=0pt]
\item We design a novel matrix encoding scheme that scales quadratically with the system size ($\mathcal{O}(n^2)$ memory), enabling Transformers to process linear systems of practical size;
\item We evaluate Algebraformer{} on a broad suite of problem instances, highlighting its robustness to ill-conditioning and noise, as well as its ability to quickly solve linear systems.

\item  We demonstrate strong generalization across two use cases: spectral interpolation problems arising in solving boundary value problems and acceleration of the Newton method (we want to remark that the Newton method is widely used in physics, for example, in inverse problems \cite{prilepko2000methods,sabatier2000past}). In the BVP experiment, we further demonstrated the adaptability of Algebraformer{} to previously unseen equations. The model was first pre-trained on a diffusion equation and then fine-tuned on a more complex one, consistently outperforming a model trained from scratch. This highlights its ability to transfer relevant knowledge and effectively generalize to datasets outside its original training distribution. The experiment concerning the Newton method is provided in Appendix \ref{sec:newton_experiment}.
\end{itemize}

While we do not aim to outperform finely tuned, task-specific numerical methods, our goal is to provide a general-purpose, plug-and-play solver that offers a compelling trade-off between accuracy, robustness, and ease of use. By sidestepping the need for handcrafted solver pipelines and extensive tuning, Algebraformer{} reduces the entry barrier to solving challenging linear systems, opening the door to broader adoption of ML-based numerical solvers in scientific and engineering workflows.

\section{Related Work}

\textbf{Learning to accelerate numerical algorithms.}
There has been a surge of interest in learning-based approaches for improving classical solvers. Several works focus on accelerating iterative methods such as Conjugate Gradient for symmetric positive definite systems \cite{li2023learning, pmlr-v202-kaneda23a, zhang2023artificial}, or GMRES-type solvers for specific applications like the Poisson equation \cite{luna2021acceleratinggmresdeeplearning}. Others focus on learned preconditioning strategies: neural networks have been used to construct preconditioners that speed up convergence \cite{pmlr-v97-greenfeld19a, pmlr-v119-luz20a, 10.5555/3540261.3541189}, or to optimize heuristics such as Jacobi and ILU variants \cite{10.1145/3441850, stanaityte2020ilu}. NeurKItt \cite{luo2024neural}, for example, employs a neural operator to predict the invariant subspace of the system matrix and accelerate solution convergence. However, these approaches are primarily designed to enhance classical numerical pipelines. In contrast, our work takes a fundamentally different perspective: we aim to learn an end-to-end solver that directly outputs the solution to a linear system, bypassing the traditional iterative or decomposition-based steps altogether. We leave for the appendix an extended related works section \cref{sec:extended_related_works}.

\section{Method}\label{sec:method}
In this section, we present the design of Algebraformer, highlighting how matrices can be efficiently encoded into a sequence-to-sequence model that can be applied out of the box to solve a wide range of linear systems.

\subsection{Algebraformer}\label{sec:algebraformer}
Transformers have become a cornerstone in many areas of machine learning, excelling at processing one-dimensional sequences of token embeddings. However, adapting them to handle inherently two-dimensional data poses significant challenges. In~\cite{charton2021linear}, the authors address this by flattening a matrix of size \(n \times n\) into a sequence, representing each matrix entry with symbolic tokens. They explore various encoding schemes: P$10$, P$1000$, P$1999$, and FP$15$, which correspond to using 5, 3, 2, and 1 token(s) per matrix entry, respectively. While this approach is innovative and has inspired further research, it also comes with notable limitations. Most critically, the memory complexity of the self-attention mechanism scales as \(\mathcal{O}(n^4)\), making it infeasible to train even on moderate-to-small matrices. Additionally, the symbolic encoding introduces significant computational overhead during inference. Finally, the method reduces numerical precision, capping accuracy at \(10^{-2}\), which may be inadequate for applications requiring finer resolution.

\begin{figure}[h]
    \centering
    \includegraphics[width=.8\linewidth]{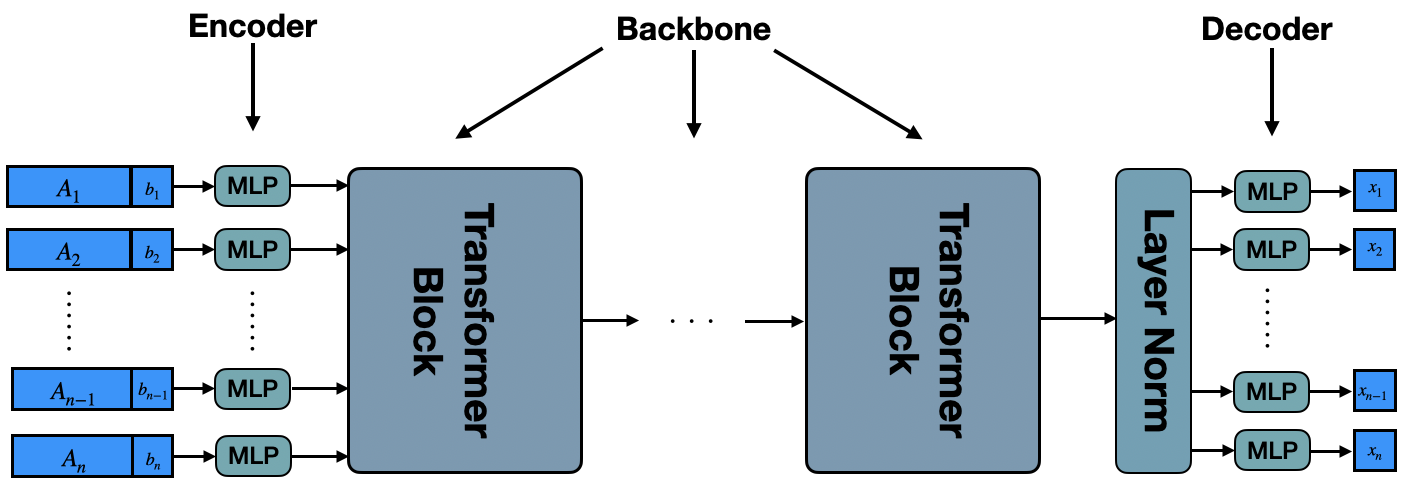}
    \caption{\textbf{Model overview.} We split the matrix $A$ into column patches, to each patch we attach one component of the vector $b$, then we embed each patch into a decoder-only Transformer backbone. At the end, we decode the output of the backbone to output the vector solution $x$}
    \label{fig:algformer_arc}
\end{figure}

Rather than flattening the matrix \(A \in \mathbb{R}^{n \times n}\) into a one-dimensional sequence of length \(n^2\), we encode the matrix and vector inputs more efficiently. Specifically, we represent each column of \(A\) separately and associate it with the corresponding entry in the right-hand side vector \(b \in \mathbb{R}^{n}\). The resulting input sequence consists of \(n\) tokens, where the \(i\)-th token is defined as \([a_i, b_i] \in \mathbb{R}^{n+1}\), with \(a_i\) denoting the \(i\)-th column of \(A\), and \(b_i\) the \(i\)-th component of \(b\). This structured encoding enables the model to process the linear system in a format that preserves its two-dimensional nature without incurring the excessive memory costs of full flattening, in fact, the requirements grow as \(\mathcal{O}(n^2)\). A graphical overview of the model architecture is provided in \cref{fig:algformer_arc}.

For the backbone, we choose a decoder-only transformer, akin to the one used in \cite{radford2019language}. The $l$-th block is represented by the equation \cref{eq:algebraformer_backbone}, where $n_1$ and $n_2$ denote the normalization layers. For normalization, we employ Layer Normalization \cite{ba2016layer}, and we use a pre-norm strategy. Additionally, the normalization layer is applied to the output of the backbone:
\begin{align}
\hat{x}_l = x_l + \mathrm{Attn}(n_1(x_l)),  \qquad 
x_{l+1} = \hat{x}_l + \mathrm{MLP}(n_2(\hat{x}_l)).\label{eq:algebraformer_backbone}
\end{align}
To train our model, we use the following loss
\begin{equation}
    \min_{\theta} \mathbb{E}\left[\mathcal{L}\left(\mathrm{Model}_{\theta}\!\left(\bar{A}\right),x\right) \right]\label{eq:loss}
\end{equation}
where $x$ is the solution of the linear system, $\bar{A}=[A,b]$, and $\mathcal{L}(x,y) = \sum_{i}(x_i-y_i)^2$ is the MSE between the ground truth solution vector and the predicted vector.

\section{Models details}

We are interested in approximating a function defined as follows
$(A, b) \mapsto x = A^{-1} b$. This mapping possesses two key properties: nonlinearity and smoothness. 
While the context differs from ours, \cite{garg2022can} demonstrates that specific classes of nonlinear smooth functions can be learned using in-context examples. Their findings show that Transformers with approximately 10M parameters can effectively approximate function classes with comparable characteristics. Another motivation to keep the model size small is that we aim to have fast inference and be competitive with standard numerical solvers. To this end, we train Algebraformer with $12$ Transformer blocks. Each block uses an embedding dimension of $256$ and 8 attention heads. In the MLP sublayers, we project the embedding dimension by a factor of 4 and use the \texttt{GELU} as activation function \cite{hendrycks2016bridging}. Our model has $9.5$M parameters.

We train two baselines: bidirectional LSTM and GRU. \cite{charton2021linear} showed that for linear algebra tasks, these RNNs perform comparably to transformers, though their study used very small matrices and did not focus on linear systems. To investigate further, we adopt an Algebraformer-like encoding for both, using 4 layers with 384-dimensional embeddings, yielding 12M parameters for the LSTM and 9M for the GRU. All three models employ a linear layer as encoder and decoder.

\section{Experiments with spectral method for BVPs}
In this section, we evaluate Algebraformer{} on linear systems arising from spectral methods for boundary value problems. We pre-train Algebraformer{} on a diffusion equation and, on this dataset, compare it with bidirectional LSTM and GRU models, as well as with standard numerical solvers (a direct method using LU decomposition with partial pivoting and least-squares solvers using SVD and QR decomposition) under noisy conditions. Furthermore, we show that Algebraformer{} can be easily fine-tuned on more complex equations in a low-data regime, specifically, a reaction-diffusion equation and an advection-diffusion equation, comparing the pre-trained model with the one trained from scratch. We want to highlight the fact that for all the equations analyzed in this dataset, the resulting matrices have a condition number on the order of \(10^5\).

We defer the introduction to boundary value problems to \cref{sec:th_bvp}, spectral methods to \cref{sec:spectral_method}, the equations used to \cref{sec:equation}, data generation to \cref{sec:data_BVP}, and training details to \cref{sec:bvp_training_details}.

\subsection{Results}

In \cref{tab:bvp_lstm_gru}, we can notice that Algebraformer{} outperforms both the LSTM and GRU baseline, on the dataset of the diffusion equation \cref{eq:darcy}.

\begin{wraptable}{r}{0.5\textwidth}
\centering
\begin{tabular}{l c}
\toprule
\textbf{Model} & \textbf{MSE} \\
\midrule
LSTM & 0.00048211 \\
GRU & 0.00031371 \\
Algebraformer & \textbf{0.00024131} \\
\bottomrule
\end{tabular}
\caption{MSE comparison on the diffusion equation.}
\label{tab:bvp_lstm_gru}
\end{wraptable}

\begin{figure}[h]
    \centering
 \includegraphics[width=1.\linewidth]{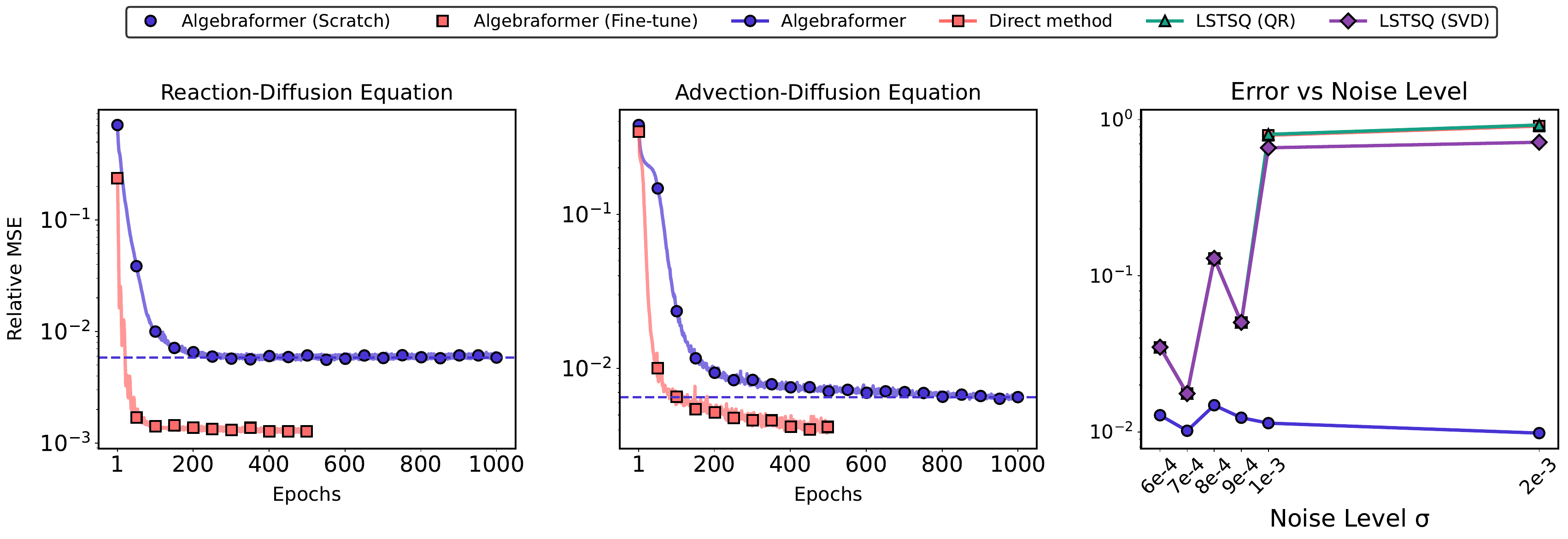}
    \caption{The two plots on the left show the relative MSE on the test set during training for \eqref{eq:darcy_absorption} and \eqref{eq:darcy_reaction}. The plot on the right displays the relative MSE on the test set with noisy data for \eqref{eq:darcy}, where we compare Algebraformer{} with $3$ different numerical methods.}
    \label{fig:plot_bvp}
\end{figure}

The first two panels on the left of \cref{fig:plot_bvp} show performance on the fine-tuning task. In both cases, the fine-tuned model outperforms the model trained from scratch, demonstrating that pretraining on the simpler diffusion equation improves the model's ability to adapt to new and more complex equations. For \cref{eq:darcy_absorption} (left), the fine-tuned model achieves an order-of-magnitude lower test error by epoch 50, a level that the scratch-trained model does not reach even after $1000$ epochs. For \cref{eq:darcy_reaction} (middle), the fine-tuned model matches the scratch-trained model’s performance at epoch $1000$ within just $100$ epochs, and continues to improve until around epoch $500$.

The right panel of \cref{fig:plot_bvp} assesses robustness to noise on the diffusion equation. Algebraformer maintains low relative MSE even with noisy test data, outperforming classical solvers such as direct methods and least-squares approaches (SVD, QR).

\section{Experiments with Newton method for nonlinear optimization}\label{sec:newton_experiment}
In this section, we focus on one more key setting where solving a linear system plays a crucial role: second-order optimization. In particular, we look at the Newton method~\cite{kelley2003solving}. One of the main drawbacks of Newton's method for the unconstrained optimization problem $\min_x f(x)$ is that, at each iteration, it requires solving the linear system $p_k= \mathcal{H}_f(x_k)^{-1} \nabla f(x_k)$ to compute the update $x_{k+1}=x_k+p_k$, from the current approximation $x_k$. Here $\mathcal{H}_f(x_k)$ is the Hessian and $\nabla f(x_k)$ is the gradient of the objective function $f$ in $x_k$. 
While the update direction $p_k$ is locally optimal, as it guarantees very fast quadratic local convergence, this procedure remains computationally prohibitive in most applications as it requires computing the Hessian matrix and solving the associated linear system at each step. Approximated or inexact Newton alternatives are popular alternatives that significantly reduce the cost per iteration by approximating the Hessian and its inverse, albeit sacrificing on the convergence speed side~\cite{dembo1982inexact,eisenstat1996choosing,kelley2003solving}. Here we show that the method can be drastically sped up by using Algebraformer{} to replace the standard linear system solver step.

As a reference problem, we consider the following minimization task:
\[
\min_{x \in \mathbb{R}^n} f(x) = \|A x - b\|_p,
\]
where $p \neq 2$, $A \in \mathbb{R}^{n \times m}$, and $b \in \mathbb{R}^m$. Choosing $p \neq 2$ ensures nonlinearity, and this is a well-studied problem in the literature~\cite{lanza2015generalized,chung2019flexible,buccini2020lp}. A detailed technical introduction to Newton's method and the problem formulation is provided in \cref{sec:newton_theory}.

As previously mentioned, constructing the Hessian and solving the corresponding linear system is costly. To address this, we propose using Algebraformer{} to take as input the current iterate $x_k$ and the problem data ($A$ and $b$), and directly predict the update direction $p_k$.

\begin{wrapfigure}{r}{0.5\textwidth}
  \centering
  \includegraphics[width=0.4\textwidth]{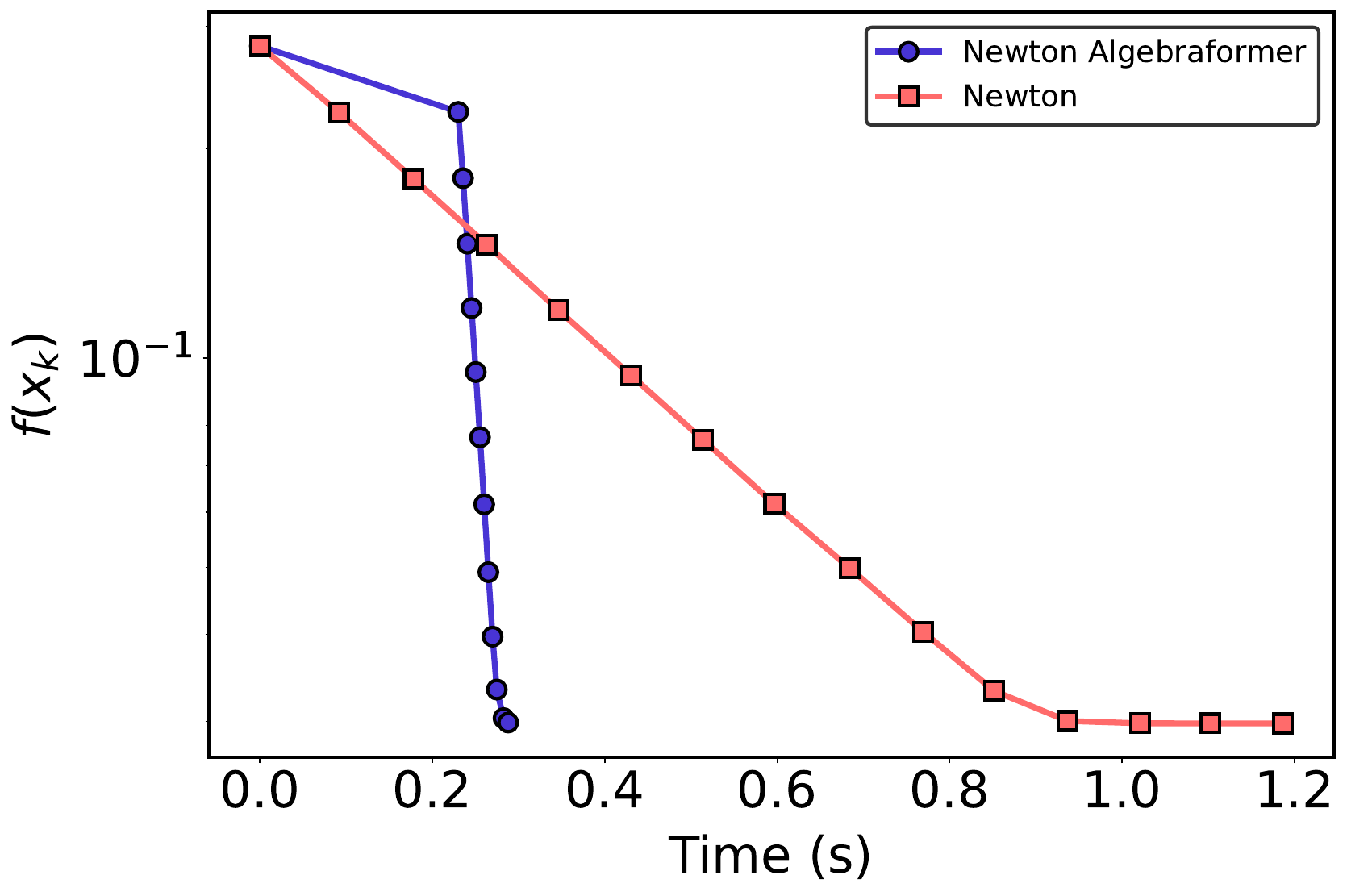}
  \caption{Time to convergence for Newton and accelerated Newton methods.}
  \label{fig:newton_acc}
\end{wrapfigure}

We conduct an experiment using matrices with $10^4$ rows and $60$ columns, with $A$ and $b$ uniformly distributed and normalized to unit norm. We conduct the experiments with $p = 6$, which is typically employed to reduce the influence of outliers. To reduce memory demand, due to the high amount of rows in the matrix $A$, instead of encoding the full matrix $A$ and vector $b$, we encode the vector $A^\top b$ concatenated with the current iterate $x_k$. We generate 1,250 Newton iteration trajectories for the training set. By \textit{trajectory}, we mean a sequence of Newton method steps \([x_0, x_1, \dots, x_k]\) until convergence, i.e., until the norm of the residual $|f(x_k)-f(x_{k+1})|$ reaches a tolerance of \(10^{-5}\). We use 125 trajectories as the test set.

For the training of our model, we use the AdamW optimizer with $\beta_1 = 0.9$ and $\beta_2 = 0.95$, without any warm-up, and apply cosine decay to the learning rate, starting from $10^{-4}$ and decaying to $10^{-5}$, and we train our model for $50$ epochs. The architecture hyperparameters remain unchanged across experiments. 

\subsection{Results}
After 50 epochs, our model can solve the linear system with a mean squared error (MSE) of $1\times 10^{-4}$ on the test set.

As shown in \cref{fig:newton_acc}, although the updates produced by our model are inexact, they enable convergence approximately four times faster than the classical Newton method. The plot also shows that both methods do converge; however, the objective function does not reach zero, as the linear system \(Ax = b\) has no exact solution. In fact, the minimum value achieved is approximately \(2 \times 10^{-2}\). Additionally, we observe that the first model evaluation takes noticeably longer than the subsequent ones. This initial latency is likely due to internal mechanisms within \texttt{PyTorch}.

\section{Conclusion}
We introduced Algebraformer, a Transformer-based architecture for solving linear systems. We show that Transformers can effectively solve linear algebra problems, and we propose a more scalable encoding framework than existing approaches in the literature. Experiments on BVP interpolation and Newton’s method (\cref{sec:newton_experiment}) demonstrate that Algebraformer offers greater robustness and speed-ups compared to standard numerical methods, highlighting its practical potential.

%
  
\newpage
\bibliographystyle{plain}
\bibliography{references}

@Inbook{Groetsch2011,
author="Groetsch, Charles",
editor="Scherzer, Otmar",
title="Linear Inverse Problems",
bookTitle="Handbook of Mathematical Methods in Imaging",
year="2011",
publisher="Springer New York",
address="New York, NY",
pages="3--41",
isbn="978-0-387-92920-0",
doi="10.1007/978-0-387-92920-0_1",
url="https://doi.org/10.1007/978-0-387-92920-0_1"
}

@book{lunardi2018interpolation,
  title={Interpolation theory},
  author={Lunardi, Alessandra},
  volume={16},
  year={2018},
  publisher={Springer}
}

@article{butcher1964implicit,
  title={Implicit runge-kutta processes},
  author={Butcher, John C},
  journal={Mathematics of computation},
  volume={18},
  number={85},
  pages={50--64},
  year={1964},
  publisher={JSTOR}
}

@book{trefethen2000spectral,
  title={Spectral methods in MATLAB},
  author={Trefethen, Lloyd N},
  year={2000},
  publisher={SIAM}
}

@book{van2003iterative,
  title={Iterative Krylov methods for large linear systems},
  author={Van der Vorst, Henk A},
  number={13},
  year={2003},
  publisher={Cambridge University Press}
}

@article{davis2016survey,
  title={A survey of direct methods for sparse linear systems},
  author={Davis, Timothy A and Rajamanickam, Sivasankaran and Sid-Lakhdar, Wissam M},
  journal={Acta Numerica},
  volume={25},
  pages={383--566},
  year={2016},
  publisher={Cambridge University Press}
}

@article{vaswani2017attention,
  title={Attention is all you need},
  author={Vaswani, Ashish and Shazeer, Noam and Parmar, Niki and Uszkoreit, Jakob and Jones, Llion and Gomez, Aidan N and Kaiser, {\L}ukasz and Polosukhin, Illia},
  journal={Advances in neural information processing systems},
  volume={30},
  year={2017}
}

@book{saad2003iterative,
  title={Iterative methods for sparse linear systems},
  author={Saad, Yousef},
  year={2003},
  publisher={SIAM}
}

@article{martinsson2020randomized,
  title={Randomized numerical linear algebra: Foundations and algorithms},
  author={Martinsson, Per-Gunnar and Tropp, Joel A},
  journal={Acta Numerica},
  volume={29},
  pages={403--572},
  year={2020},
  publisher={Cambridge University Press}
}

@article{strohmer2009randomized,
  title={A randomized Kaczmarz algorithm with exponential convergence},
  author={Strohmer, Thomas and Vershynin, Roman},
  journal={Journal of Fourier Analysis and Applications},
  volume={15},
  number={2},
  pages={262--278},
  year={2009},
  publisher={Springer}
}

@article{
charton2021linear,
title={Linear algebra with transformers},
author={Francois Charton},
journal={Transactions on Machine Learning Research},
issn={2835-8856},
year={2022},
url={https://openreview.net/forum?id=Hp4g7FAXXG},
note={}
}

@article{herde2024poseidon,
  title={Poseidon: Efficient foundation models for pdes},
  author={Herde, Maximilian and Raonic, Bogdan and Rohner, Tobias and K{\"a}ppeli, Roger and Molinaro, Roberto and de B{\'e}zenac, Emmanuel and Mishra, Siddhartha},
  journal={Advances in Neural Information Processing Systems},
  volume={37},
  pages={72525--72624},
  year={2024}
}

@article{mccabe2023multiple,
  title={Multiple physics pretraining for physical surrogate models},
  author={McCabe, Michael and Blancard, Bruno R{\'e}galdo-Saint and Parker, Liam Holden and Ohana, Ruben and Cranmer, Miles and Bietti, Alberto and Eickenberg, Michael and Golkar, Siavash and Krawezik, Geraud and Lanusse, Francois and others},
  journal={arXiv preprint arXiv:2310.02994},
  year={2023}
}

@article{tikhonov1963solution,
  title={Solution of incorrectly formulated problems and the regularization method.},
  author={Tikhonov, Andrei N},
  journal={Sov Dok},
  volume={4},
  pages={1035--1038},
  year={1963}
}

@article{Phillips_ref,
author = {Phillips, David L.},
title = {A Technique for the Numerical Solution of Certain Integral Equations of the First Kind},
year = {1962},
issue_date = {Jan. 1962},
publisher = {Association for Computing Machinery},
address = {New York, NY, USA},
volume = {9},
number = {1},
issn = {0004-5411},
url = {https://doi.org/10.1145/321105.321114},
doi = {10.1145/321105.321114},
journal = {J. ACM},
month = jan,
pages = {84–97},
numpages = {14}
}

@book{chen2005matrix,
  title={Matrix preconditioning techniques and applications},
  author={Chen, Ke},
  number={19},
  year={2005},
  publisher={Cambridge University Press}
}

@article{lample2019deep,
  title={Deep learning for symbolic mathematics},
  author={Lample, Guillaume and Charton, Fran{\c{c}}ois},
  journal={arXiv preprint arXiv:1912.01412},
  year={2019}
}

@article{polu2020generative,
  title={Generative language modeling for automated theorem proving},
  author={Polu, Stanislas and Sutskever, Ilya},
  journal={arXiv preprint arXiv:2009.03393},
  year={2020}
}

@article{hahn2020teaching,
  title={Teaching temporal logics to neural networks},
  author={Hahn, Christopher and Schmitt, Frederik and Kreber, Jens U and Rabe, Markus N and Finkbeiner, Bernd},
  journal={arXiv preprint arXiv:2003.04218},
  year={2020}
}

@article{shi2021transformer,
  title={Transformer-based machine learning for fast SAT solvers and logic synthesis},
  author={Shi, Feng and Lee, Chonghan and Bashar, Mohammad Khairul and Shukla, Nikhil and Zhu, Song-Chun and Narayanan, Vijaykrishnan},
  journal={arXiv preprint arXiv:2107.07116},
  year={2021}
}

@inproceedings{biggio2021neural,
  title={Neural symbolic regression that scales},
  author={Biggio, Luca and Bendinelli, Tommaso and Neitz, Alexander and Lucchi, Aurelien and Parascandolo, Giambattista},
  booktitle={International Conference on Machine Learning},
  pages={936--945},
  year={2021},
  organization={Pmlr}
}

@article{charton2020learning,
  title={Learning advanced mathematical computations from examples},
  author={Charton, Fran{\c{c}}ois and Hayat, Amaury and Lample, Guillaume},
  journal={arXiv preprint arXiv:2006.06462},
  year={2020}
}

@article{davies2021advancing,
  title={Advancing mathematics by guiding human intuition with AI},
  author={Davies, Alex and Veli{\v{c}}kovi{\'c}, Petar and Buesing, Lars and Blackwell, Sam and Zheng, Daniel and Toma{\v{s}}ev, Nenad and Tanburn, Richard and Battaglia, Peter and Blundell, Charles and Juh{\'a}sz, Andr{\'a}s and others},
  journal={Nature},
  volume={600},
  number={7887},
  pages={70--74},
  year={2021},
  publisher={Nature Publishing Group UK London}
}

@article{frieder2023mathematical,
  title={Mathematical capabilities of chatgpt},
  author={Frieder, Simon and Pinchetti, Luca and Griffiths, Ryan-Rhys and Salvatori, Tommaso and Lukasiewicz, Thomas and Petersen, Philipp and Berner, Julius},
  journal={Advances in neural information processing systems},
  volume={36},
  pages={27699--27744},
  year={2023}
}

@article{frieder2023large,
  title={Large language models for mathematicians},
  author={Frieder, Simon and Berner, Julius and Petersen, Philipp and Lukasiewicz, Thomas},
  journal={arXiv preprint arXiv:2312.04556},
  year={2023}
}

@article{guo2025deepseek,
  title={Deepseek-r1: Incentivizing reasoning capability in llms via reinforcement learning},
  author={Guo, Daya and Yang, Dejian and Zhang, Haowei and Song, Junxiao and Zhang, Ruoyu and Xu, Runxin and Zhu, Qihao and Ma, Shirong and Wang, Peiyi and Bi, Xiao and others},
  journal={arXiv preprint arXiv:2501.12948},
  year={2025}
}

@article{zhang2023can,
  title={Can transformers learn to solve problems recursively?},
  author={Zhang, Shizhuo Dylan and Tigges, Curt and Biderman, Stella and Raginsky, Maxim and Ringer, Talia},
  journal={arXiv preprint arXiv:2305.14699},
  year={2023}
}

@article{garg2022can,
  title={What can transformers learn in-context? a case study of simple function classes},
  author={Garg, Shivam and Tsipras, Dimitris and Liang, Percy S and Valiant, Gregory},
  journal={Advances in Neural Information Processing Systems},
  volume={35},
  pages={30583--30598},
  year={2022}
}

@article{yang2023looped,
  title={Looped transformers are better at learning learning algorithms},
  author={Yang, Liu and Lee, Kangwook and Nowak, Robert and Papailiopoulos, Dimitris},
  journal={arXiv preprint arXiv:2311.12424},
  year={2023}
}

@article{kantamneni2025language,
  title={Language Models Use Trigonometry to Do Addition},
  author={Kantamneni, Subhash and Tegmark, Max},
  journal={arXiv preprint arXiv:2502.00873},
  year={2025}
}

@article{li2020fourier,
  title={Fourier neural operator for parametric partial differential equations},
  author={Li, Zongyi and Kovachki, Nikola and Azizzadenesheli, Kamyar and Liu, Burigede and Bhattacharya, Kaushik and Stuart, Andrew and Anandkumar, Anima},
  journal={arXiv preprint arXiv:2010.08895},
  year={2020}
}

@article{tran2021factorized,
  title={Factorized fourier neural operators},
  author={Tran, Alasdair and Mathews, Alexander and Xie, Lexing and Ong, Cheng Soon},
  journal={arXiv preprint arXiv:2111.13802},
  year={2021}
}

@article{wen2022u,
  title={U-FNO—An enhanced Fourier neural operator-based deep-learning model for multiphase flow},
  author={Wen, Gege and Li, Zongyi and Azizzadenesheli, Kamyar and Anandkumar, Anima and Benson, Sally M},
  journal={Advances in Water Resources},
  volume={163},
  pages={104180},
  year={2022},
  publisher={Elsevier}
}

@article{azizzadenesheli2024neural,
  title={Neural operators for accelerating scientific simulations and design},
  author={Azizzadenesheli, Kamyar and Kovachki, Nikola and Li, Zongyi and Liu-Schiaffini, Miguel and Kossaifi, Jean and Anandkumar, Anima},
  journal={Nature Reviews Physics},
  volume={6},
  number={5},
  pages={320--328},
  year={2024},
  publisher={Nature Publishing Group UK London}
}

@incollection{goswami2023physics,
  title={Physics-informed deep neural operator networks},
  author={Goswami, Somdatta and Bora, Aniruddha and Yu, Yue and Karniadakis, George Em},
  booktitle={Machine learning in modeling and simulation: methods and applications},
  pages={219--254},
  year={2023},
  publisher={Springer}
}

@inproceedings{hao2023gnot,
  title={Gnot: A general neural operator transformer for operator learning},
  author={Hao, Zhongkai and Wang, Zhengyi and Su, Hang and Ying, Chengyang and Dong, Yinpeng and Liu, Songming and Cheng, Ze and Song, Jian and Zhu, Jun},
  booktitle={International Conference on Machine Learning},
  pages={12556--12569},
  year={2023},
  organization={PMLR}
}

@inproceedings{raonic2023convolutional,
  title={Convolutional neural operators},
  author={Raonic, Bogdan and Molinaro, Roberto and Rohner, Tobias and Mishra, Siddhartha and de Bezenac, Emmanuel},
  booktitle={ICLR 2023 Workshop on Physics for Machine Learning},
  year={2023}
}

@article{brandstetter2022message,
  title={Message passing neural PDE solvers},
  author={Brandstetter, Johannes and Worrall, Daniel and Welling, Max},
  journal={arXiv preprint arXiv:2202.03376},
  year={2022}
}

@article{gupta2022towards,
  title={Towards multi-spatiotemporal-scale generalized pde modeling},
  author={Gupta, Jayesh K and Brandstetter, Johannes},
  journal={arXiv preprint arXiv:2209.15616},
  year={2022}
}

@article{ye2024pdeformer,
  title={Pdeformer: Towards a foundation model for one-dimensional partial differential equations},
  author={Ye, Zhanhong and Huang, Xiang and Chen, Leheng and Liu, Hongsheng and Wang, Zidong and Dong, Bin},
  journal={arXiv preprint arXiv:2402.12652},
  year={2024}
}

@inproceedings{towards23,
author = {Subramanian, Shashank and Harrington, Peter and Keutzer, Kurt and Bhimji, Wahid and Morozov, Dmitriy and Mahoney, Michael W. and Gholami, Amir},
title = {Towards foundation models for scientific machine learning: characterizing scaling and transfer behavior},
year = {2023},
publisher = {Curran Associates Inc.},
address = {Red Hook, NY, USA},
booktitle = {Proceedings of the 37th International Conference on Neural Information Processing Systems},
articleno = {3119},
numpages = {21},
location = {New Orleans, LA, USA},
series = {NIPS '23}
}

@misc{veličković2024amplifyinghumanperformancecombinatorial,
      title={Amplifying human performance in combinatorial competitive programming}, 
      author={Petar Veličković and Alex Vitvitskyi and Larisa Markeeva and Borja Ibarz and Lars Buesing and Matej Balog and Alexander Novikov},
      year={2024},
      eprint={2411.19744},
      archivePrefix={arXiv},
      primaryClass={cs.LG},
      url={https://arxiv.org/abs/2411.19744}, 
}

@misc{velikovi2019neural,
    title={Neural Execution of Graph Algorithms},
    author={Petar Veličković and Rex Ying and Matilde Padovano and Raia Hadsell and Charles Blundell},
    year={2019},
    eprint={1910.10593},
    archivePrefix={arXiv},
    primaryClass={stat.ML}
}

@InProceedings{pmlr-v162-velickovic22a,
  title = 	 {The {CLRS} Algorithmic Reasoning Benchmark},
  author =       {Veli{\v{c}}kovi{\'c}, Petar and Badia, Adri{\`a} Puigdom{\`e}nech and Budden, David and Pascanu, Razvan and Banino, Andrea and Dashevskiy, Misha and Hadsell, Raia and Blundell, Charles},
  booktitle = 	 {Proceedings of the 39th International Conference on Machine Learning},
  pages = 	 {22084--22102},
  year = 	 {2022},
  editor = 	 {Chaudhuri, Kamalika and Jegelka, Stefanie and Song, Le and Szepesvari, Csaba and Niu, Gang and Sabato, Sivan},
  volume = 	 {162},
  series = 	 {Proceedings of Machine Learning Research},
  month = 	 {17--23 Jul},
  publisher =    {PMLR},
  url = 	 {https://proceedings.mlr.press/v162/velickovic22a.html}
}

@inproceedings{
georgiev2024deep,
title={Deep Equilibrium Algorithmic Reasoning},
author={Dobrik Georgiev Georgiev and JJ Wilson and Davide Buffelli and Pietro Lio},
booktitle={The Thirty-eighth Annual Conference on Neural Information Processing Systems},
year={2024},
url={https://openreview.net/forum?id=SuLxkxCENa}
}

@misc{rec,
title={Recursive Reasoning with Neural Networks},
author={Jonas J{\"u}r{\ss} and Dulhan Hansaja Jayalath},
year={2023},
url={https://openreview.net/forum?id=TS8l4VS7_BK}
}

@article{DBLP/corr/abs-2409-07154,
  author       = {Kaijia Xu and
                  Petar Velickovic},
  title        = {Recurrent Aggregators in Neural Algorithmic Reasoning},
  journal      = {CoRR},
  volume       = {abs/2409.07154},
  year         = {2024},
  url          = {https://doi.org/10.48550/arXiv.2409.07154},
  doi          = {10.48550/ARXIV.2409.07154},
  eprinttype    = {arXiv},
  eprint       = {2409.07154},
  bibsource    = {dblp computer science bibliography, https://dblp.org}
}

@misc{
bounsi2025transformers,
title={Transformers meet Neural Algorithmic Reasoners},
author={Wilfried Bounsi and Borja Ibarz and Andrew Joseph Dudzik and Jessica B Hamrick and Larisa Markeeva and Alex Vitvitskyi and Razvan Pascanu and Petar Veli{\v{c}}kovi{\'c}},
year={2025},
url={https://openreview.net/forum?id=fk4czNKXPC}
}

@inproceedings{
deac2021neural,
title={Neural Algorithmic Reasoners are Implicit Planners},
author={Andreea Deac and Petar Veli{\v{c}}kovi{\'c} and Ognjen Milinkovi{\'c} and Pierre-Luc Bacon and Jian Tang and Mladen Nikolic},
booktitle={Advances in Neural Information Processing Systems},
editor={A. Beygelzimer and Y. Dauphin and P. Liang and J. Wortman Vaughan},
year={2021},
url={https://openreview.net/forum?id=K5YKjaMjbja}
}

@misc{li2023learning,
    title={Learning Preconditioner for Conjugate Gradient PDE Solvers},
    author={Yichen Li and Peter Yichen Chen and Tao Du and Wojciech Matusik},
    year={2023},
    eprint={2305.16432},
    archivePrefix={arXiv},
    primaryClass={math.NA}
}

@InProceedings{pmlr-v202-kaneda23a,
  title = 	 {A Deep Conjugate Direction Method for Iteratively Solving Linear Systems},
  author =       {Kaneda, Ayano and Akar, Osman and Chen, Jingyu and Kala, Victoria Alicia Trevino and Hyde, David and Teran, Joseph},
  booktitle = 	 {Proceedings of the 40th International Conference on Machine Learning},
  pages = 	 {15720--15736},
  year = 	 {2023},
  editor = 	 {Krause, Andreas and Brunskill, Emma and Cho, Kyunghyun and Engelhardt, Barbara and Sabato, Sivan and Scarlett, Jonathan},
  volume = 	 {202},
  series = 	 {Proceedings of Machine Learning Research},
  month = 	 {23--29 Jul},
  publisher =    {PMLR},
  url = 	 {https://proceedings.mlr.press/v202/kaneda23a.html}
}

@misc{zhang2023artificial,
    title={Artificial Intelligence for Science in Quantum, Atomistic, and Continuum Systems},
    author={Xuan Zhang and Limei Wang and Jacob Helwig and Youzhi Luo and Cong Fu and Yaochen Xie and Meng Liu and Yuchao Lin and Zhao Xu and Keqiang Yan and Keir Adams and Maurice Weiler and Xiner Li and Tianfan Fu and Yucheng Wang and Alex Strasser and Haiyang Yu and YuQing Xie and Xiang Fu and Shenglong Xu and Yi Liu and Yuanqi Du and Alexandra Saxton and Hongyi Ling and Hannah Lawrence and Hannes Stärk and Shurui Gui and Carl Edwards and Nicholas Gao and Adriana Ladera and Tailin Wu and Elyssa F. Hofgard and Aria Mansouri Tehrani and Rui Wang and Ameya Daigavane and Montgomery Bohde and Jerry Kurtin and Qian Huang and Tuong Phung and Minkai Xu and Chaitanya K. Joshi and Simon V. Mathis and Kamyar Azizzadenesheli and Ada Fang and Alán Aspuru-Guzik and Erik Bekkers and Michael Bronstein and Marinka Zitnik and Anima Anandkumar and Stefano Ermon and Pietro Liò and Rose Yu and Stephan Günnemann and Jure Leskovec and Heng Ji and Jimeng Sun and Regina Barzilay and Tommi Jaakkola and Connor W. Coley and Xiaoning Qian and Xiaofeng Qian and Tess Smidt and Shuiwang Ji},
    year={2023},
    eprint={2307.08423},
    archivePrefix={arXiv},
    primaryClass={cs.LG}
}

@misc{luna2021acceleratinggmresdeeplearning,
      title={Accelerating GMRES with Deep Learning in Real-Time}, 
      author={Kevin Luna and Katherine Klymko and Johannes P. Blaschke},
      year={2021},
      eprint={2103.10975},
      archivePrefix={arXiv},
      primaryClass={physics.comp-ph},
      url={https://arxiv.org/abs/2103.10975}, 
}

@inproceedings{
luo2024neural,
title={Neural Krylov Iteration for Accelerating Linear System Solving},
author={Jian Luo and Jie Wang and Hong Wang and huanshuo dong and Zijie Geng and Hanzhu Chen and Yufei Kuang},
booktitle={The Thirty-eighth Annual Conference on Neural Information Processing Systems},
year={2024},
url={https://openreview.net/forum?id=cqfE9eYMdP}
}

@article{radford2019language,
  title={Language models are unsupervised multitask learners},
  author={Radford, Alec and Wu, Jeffrey and Child, Rewon and Luan, David and Amodei, Dario and Sutskever, Ilya and others}
}

@article{ba2016layer,
  title={Layer normalization},
  author={Ba, Jimmy Lei and Kiros, Jamie Ryan and Hinton, Geoffrey E},
  journal={arXiv preprint arXiv:1607.06450},
  year={2016}
}

@article{hendrycks2016bridging,
  title={Bridging Nonlinearities and Stochastic Regularizers with Gaussian Error Linear Units},
  author={Hendrycks, Dan and Gimpel, Kevin},
  year={2016}
}

@InProceedings{pmlr-v97-greenfeld19a,
  title = 	 {Learning to Optimize Multigrid {PDE} Solvers},
  author =       {Greenfeld, Daniel and Galun, Meirav and Basri, Ronen and Yavneh, Irad and Kimmel, Ron},
  booktitle = 	 {Proceedings of the 36th International Conference on Machine Learning},
  pages = 	 {2415--2423},
  year = 	 {2019},
  editor = 	 {Chaudhuri, Kamalika and Salakhutdinov, Ruslan},
  volume = 	 {97},
  series = 	 {Proceedings of Machine Learning Research},
  month = 	 {09--15 Jun},
  publisher =    {PMLR},
  url = 	 {https://proceedings.mlr.press/v97/greenfeld19a.html}
}

@InProceedings{pmlr-v119-luz20a,
  title = 	 {Learning Algebraic Multigrid Using Graph Neural Networks},
  author =       {Luz, Ilay and Galun, Meirav and Maron, Haggai and Basri, Ronen and Yavneh, Irad},
  booktitle = 	 {Proceedings of the 37th International Conference on Machine Learning},
  pages = 	 {6489--6499},
  year = 	 {2020},
  editor = 	 {III, Hal Daumé and Singh, Aarti},
  volume = 	 {119},
  series = 	 {Proceedings of Machine Learning Research},
  month = 	 {13--18 Jul},
  publisher =    {PMLR},
  url = 	 {https://proceedings.mlr.press/v119/luz20a.html}
}

@inproceedings{10.5555/3540261.3541189,
author = {Taghibakhshi, Ali and MacLachlan, Scott and Olson, Luke and West, Matthew},
title = {Optimization-based algebraic multigrid coarsening using reinforcement learning},
year = {2021},
isbn = {9781713845393},
publisher = {Curran Associates Inc.},
address = {Red Hook, NY, USA},
booktitle = {Proceedings of the 35th International Conference on Neural Information Processing Systems},
articleno = {928},
numpages = {12},
series = {NIPS '21}
}

@article{10.1145/3441850,
author = {Flegar, Goran and Anzt, Hartwig and Cojean, Terry and Quintana-Ort\'{\i}, Enrique S.},
title = {Adaptive Precision Block-Jacobi for High Performance Preconditioning in the Ginkgo Linear Algebra Software},
year = {2021},
issue_date = {June 2021},
publisher = {Association for Computing Machinery},
address = {New York, NY, USA},
volume = {47},
number = {2},
issn = {0098-3500},
url = {https://doi.org/10.1145/3441850},
doi = {10.1145/3441850},
journal = {ACM Trans. Math. Softw.},
month = apr,
articleno = {14},
numpages = {28}
}

@phdthesis{stanaityte2020ilu,
  title={ILU and Machine Learning Based Preconditioning For The Discretized Incompressible Navier-Stokes Equations},
  author={Stanaityte, Rita},
  year={2020},
  school={University of Houston}
}

@inproceedings{shumaylovweakly,
  title={Weakly Convex Regularisers for Inverse Problems: Convergence of Critical Points and Primal-Dual Optimisation},
  author={Shumaylov, Zakhar and Budd, Jeremy and Mukherjee, Subhadip and Sch{\"o}nlieb, Carola-Bibiane},
  booktitle={Forty-first International Conference on Machine Learning}
}

@article{Total_Generalized_Variatio,
author = {Bredies, Kristian and Kunisch, Karl and Pock, Thomas},
title = {Total Generalized Variation},
journal = {SIAM Journal on Imaging Sciences},
volume = {3},
number = {3},
pages = {492-526},
year = {2010},
doi = {10.1137/090769521},

URL = { 
    
        https://doi.org/10.1137/090769521
    
    

},
eprint = { 
    
        https://doi.org/10.1137/090769521
    
    

}
}

@article{rudin1992nonlinear,
  title={Nonlinear total variation based noise removal algorithms},
  author={Rudin, Leonid I and Osher, Stanley and Fatemi, Emad},
  journal={Physica D: nonlinear phenomena},
  volume={60},
  number={1-4},
  pages={259--268},
  year={1992},
  publisher={Elsevier}
}

@article{Gilboa_Guy,
author = {Gilboa, Guy and Osher, Stanley},
title = {Nonlocal Operators with Applications to Image Processing},
journal = {Multiscale Modeling \& Simulation},
volume = {7},
number = {3},
pages = {1005-1028},
year = {2009},
doi = {10.1137/070698592},

URL = { 
    
        https://doi.org/10.1137/070698592
    
    

},
eprint = { 
    
        https://doi.org/10.1137/070698592
    
    

}
}

@article{saksman2009discretization,
  title={Discretization-invariant Bayesian inversion and Besov space priors},
  author={Saksman, Matti Lassas and Siltanen, Samuli and others},
  journal={arXiv preprint arXiv:0901.4220},
  year={2009}
}

@article{davoli2019adaptive,
  title={Adaptive image processing: first order PDE constraint regularizers and a bilevel training scheme},
  author={Davoli, Elisa and Fonseca, Irene and Liu, Pan},
  journal={arXiv preprint arXiv:1902.01122},
  year={2019}
}

@article{meijerink1977iterative,
  title={An iterative solution method for linear systems of which the coefficient matrix is a symmetric M-matrix},
  author={Meijerink, J.A. and Van der Vorst, H.A.},
  journal={Mathematics of Computation},
  volume={31},
  number={137},
  pages={148--162},
  year={1977},
  publisher={JSTOR}
}

@article{benzi2002preconditioning,
  title={Preconditioning techniques for large linear systems: a survey},
  author={Benzi, Michele},
  journal={Journal of Computational Physics},
  volume={182},
  number={2},
  pages={418--477},
  year={2002},
  publisher={Elsevier}
}

@article{brandt1977multi,
  title={Multi-level adaptive solutions to boundary-value problems},
  author={Brandt, Achi},
  journal={Mathematics of Computation},
  volume={31},
  number={138},
  pages={333--390},
  year={1977},
  publisher={JSTOR}
}

@book{kelley2003solving,
  title={Solving Nonlinear Equations with Newton's Method},
  author={Kelley, C. T.},
  year={2003},
  publisher={Society for Industrial and Applied Mathematics},
  address={Philadelphia},
  doi={10.1137/1.9780898718898}
}

@article{waltz2006interior,
  title={An interior algorithm for nonlinear optimization that combines line search and trust region steps},
  author={Waltz, Richard A and Morales, Jorge L and Nocedal, Jorge and Orban, Dominique},
  journal={Mathematical Programming},
  volume={107},
  number={3},
  pages={391--408},
  year={2006},
  publisher={Springer},
  doi={10.1007/s10107-005-0598-y}
}

@article{lin2008trust,
  title={Trust region Newton methods for large-scale logistic regression},
  author={Lin, Chih-Jen and Weng, Rong-En and Keerthi, S Sathiya},
  journal={Journal of Machine Learning Research},
  volume={9},
  pages={627--650},
  year={2008}
}

@article{dembo1982inexact,
  title={Inexact Newton methods},
  author={Dembo, Ron S and Eisenstat, Stanley C and Steihaug, Trond},
  journal={SIAM Journal on Numerical Analysis},
  volume={19},
  number={2},
  pages={400--408},
  year={1982},
  publisher={SIAM},
  doi={10.1137/0719032}
}

@article{eisenstat1996choosing,
  title={Choosing the forcing terms in an inexact Newton method},
  author={Eisenstat, Stanley C and Walker, Homer F},
  journal={SIAM Journal on Scientific Computing},
  volume={17},
  number={1},
  pages={16--32},
  year={1996},
  publisher={SIAM},
  doi={10.1137/0917058}
}

@article{lanza2015generalized,
  title={A generalized Krylov subspace method for $\backslash$ell\_p-$\backslash$ell\_q minimization},
  author={Lanza, Alessandro and Morigi, Serena and Reichel, Lothar and Sgallari, Fiorella},
  journal={SIAM Journal on Scientific Computing},
  volume={37},
  number={5},
  pages={S30--S50},
  year={2015},
  publisher={SIAM}
}

@article{chung2019flexible,
  title={Flexible Krylov methods for $\backslash$ell\_p regularization},
  author={Chung, Julianne and Gazzola, Silvia},
  journal={SIAM Journal on Scientific Computing},
  volume={41},
  number={5},
  pages={S149--S171},
  year={2019},
  publisher={SIAM}
}

@article{buccini2020lp,
  title={An lp-lq minimization method with cross-validation for the restoration of impulse noise contaminated images},
  author={Buccini, Alessandro and Reichel, Lothar},
  journal={Journal of Computational and Applied Mathematics},
  volume={375},
  pages={112824},
  year={2020},
  publisher={Elsevier}
}

@book{prilepko2000methods,
  title={Methods for solving inverse problems in mathematical physics},
  author={Prilepko, Aleksey I and Orlovsky, Dmitry G and Vasin, Igor A and others},
  year={2000},
  publisher={CRC Press}
}

@article{sabatier2000past,
  title={Past and future of inverse problems},
  author={Sabatier, Pierre C},
  journal={Journal of Mathematical Physics},
  volume={41},
  number={6},
  pages={4082--4124},
  year={2000},
  publisher={American Institute of Physics}
}

\newpage

\appendix

\section{Extended related works}\label{sec:extended_related_works}

\textbf{Transformers for mathematical tasks and algorithmic reasoning.} 
Transformers have been extensively explored for symbolic manipulation and mathematical reasoning tasks, including equation solving and symbolic integration \cite{lample2019deep, polu2020generative, hahn2020teaching, shi2021Transformer, biggio2021neural, charton2020learning}. More recently, large language models (LLMs) have demonstrated remarkable capabilities in arithmetic and logical reasoning \cite{davies2021advancing, frieder2023mathematical, frieder2023large, guo2025deepseek}. Studies have begun to shed light on the internal mechanisms behind this behavior: for example, \cite{kantamneni2025language} suggests that LLMs internally use trigonometric structures to perform operations like addition. In \cite{zhang2023can}, it was shown that Transformers can emulate structurally recursive functions from input-output data, while \cite{garg2022can, yang2023looped} examined how Transformers trained from scratch generalize to function classes in context. Another active line of work investigates how Transformers can emulate classical algorithms, leveraging recursion \cite{georgiev2024deep, rec}, or integrating graph-based inductive biases \cite{bounsi2025Transformers}. These methods have been applied in diverse settings, from competitive programming tasks \cite{veličković2024amplifyinghumanperformancecombinatorial} to fixed-point iteration problems in reinforcement learning \cite{deac2021neural}.

\textbf{Neural PDE solvers.}
A growing body of work applies neural networks to solve partial differential equations (PDEs). Among the most prominent frameworks are neural operators \cite{azizzadenesheli2024neural}, which aim to learn mappings between function spaces in a resolution-invariant fashion. Notable examples include the Fourier Neural Operator \cite{li2020fourier}, Factorized FNO \cite{tran2021factorized}, U-Net FNO \cite{wen2022u}, Physics-Informed Neural Operator \cite{goswami2023physics}, GNOT \cite{hao2023gnot}, and Convolutional Neural Operator \cite{raonic2023convolutional}. These models are particularly suited for modeling the infinite-dimensional solution operators of PDEs. Beyond operator learning, architectures like message-passing neural networks (MPNNs) and U-Nets have also proven effective in discretized PDE settings \cite{brandstetter2022message, gupta2022towards}. More recently, foundation models for PDEs have emerged \cite{mccabe2023multiple, herde2024poseidon, ye2024pdeformer, towards23}, aiming to generalize across families of equations that arise in diverse physical domains.

\textbf{Learning to accelerate numerical algorithms.}
There has been a surge of interest in learning-based approaches for improving classical solvers. Several works focus on accelerating iterative methods such as Conjugate Gradient for symmetric positive definite systems \cite{li2023learning, pmlr-v202-kaneda23a, zhang2023artificial}, or GMRES-type solvers for specific applications like the Poisson equation \cite{luna2021acceleratinggmresdeeplearning}. Others focus on learned preconditioning strategies: neural networks have been used to construct preconditioners that speed up convergence \cite{pmlr-v97-greenfeld19a, pmlr-v119-luz20a, 10.5555/3540261.3541189}, or to optimize heuristics such as Jacobi and ILU variants \cite{10.1145/3441850, stanaityte2020ilu}. NeurKItt \cite{luo2024neural}, for example, employs a neural operator to predict the invariant subspace of the system matrix and accelerate solution convergence.

\section{Boundary value problems and spectral methods}\label{sec:th_bvp}
A boundary value problem is given by a differential equation of the form
\begin{equation} \label{eq:BVP}
\left\{
\begin{aligned}
\mathcal{L}\,u(x) &= f(x), \qquad x \in \Omega,\\
B[u(x)] &= 0, \qquad x \in \partial \Omega,
\end{aligned}
\right.
\end{equation}
where \(\Omega \subset \mathbb{R}^d\) is a bounded domain with boundary \(\partial \Omega\). The differential operator \(\mathcal{L}: \mathcal{X} \to \mathcal{Y}\) acts between suitable functional spaces, with \(u: \Omega \to \mathbb{R}\) denoting the unknown solution and \(f: \Omega \to \mathbb{R}\) the prescribed source term. The boundary operator \(B\) imposes conditions on \(\partial \Omega\).

Spectral methods are a widely used class of numerical techniques known for their exponential convergence and global, basis-coefficient representation of solutions \cite{trefethen2000spectral}. In contrast, finite difference schemes (and related discretizations) typically offer only polynomial convergence and provide local approximations. However, finite difference methods often yield structured matrices, which are generally easier to handle than the dense matrices that arise in spectral methods. These observations motivate our approach: our model is agnostic to the matrix structure and can efficiently solve the resulting linear system, while preserving the key advantages of spectral methods, namely, exponential convergence and a polynomial representation of the solution.

Among spectral methods, a common choice involves Chebyshev differentiation matrices \cite{trefethen2000spectral} (see \cref{sec:spectral_method}). This formulation is particularly attractive, as it interprets the model as an oracle that outputs the node of a polynomial. Polynomials are mathematically tractable and interpretable objects, an essential advantage for applications in physics.

\section{Spectral Method and chebyshev differentiation matrix}\label{sec:spectral_method}
This section is based on the book \cite{trefethen2000spectral}. For a smooth function $u:[-1,1]\!\to\!\mathbb{R}$ we employ a Chebyshev spectral interpolant  
\[
u_N(x)=\sum_{k=0}^{N} a_k T_k(x), \qquad  T_k(\cos\theta)=\cos(k\theta),
\]
sampling at the \emph{Chebyshev--Gauss--Lobatto} nodes  
\[
x_j=\cos\!\Bigl(\tfrac{\pi j}{N}\Bigr),\qquad j=0,\dots,N,
\]
with data $u_j=u(x_j)$.  The cardinal functions $\ell_j(x)$ satisfy $\ell_j(x_m)=\delta_{jm}$, so the derivative of the interpolant is  
\[
u_N'(x)=\sum_{j=0}^{N}u_j\,\ell_j'(x).
\]
Evaluating at every node $x_i$ yields the \emph{Chebyshev differentiation matrix} $D\in\mathbb{R}^{(N+1)\times(N+1)}$ through  
\[
u_N'(x_i)=\sum_{j=0}^{N} D_{ij} u_j,\qquad D_{ij}:=\ell_j'(x_i).
\]
\textbf{Closed‐form entries}  Let $c_0=c_N=2$ and $c_j=1$ for $1\le j\le N-1$. Then  
\[
D_{ij}=
\begin{cases}
\displaystyle\frac{c_i}{c_j}\,\frac{(-1)^{i+j}}{x_i-x_j}, & i\neq j,\\[6pt]
\displaystyle-\frac{x_i}{2(1-x_i^2)},                  & 1\le i\le N-1,\\[6pt]
\displaystyle\frac{2N^{2}+1}{6},                       & i=j=0,\\[6pt]
-\displaystyle\frac{2N^{2}+1}{6},                      & i=j=N.
\end{cases}
\]

\medskip
\textbf{Spectral convergence theorem (Bernstein–Jackson–Clenshaw–Curtis).}  
If $u$ is analytic in a Bernstein ellipse $E_\rho$ ($\rho>1$) with foci $\pm1$, then for any integer $m\!\ge\!0$
\begin{equation}\label{eq:error:bound}
    \max_{x\in[-1,1]}\bigl|u^{(m)}(x)-u_N^{(m)}(x)\bigr|
\;\le\;
C_m\,\dfrac{\rho^{-N}}{(\rho-1)^{m}},
\end{equation}
where $C_m$ depends on $u$ and $\rho$.

\section{Equations}\label{sec:equation}
We consider a sequence of three differential equations on the one-dimensional domain \([0, 7.5]\) with Dirichlet boundary conditions, aiming to model diffusion-dominated processes and extend to more complex physical phenomena.

We first train our model on a classical second-order elliptic equation describing diffusion in a heterogeneous medium:
\begin{equation}\label{eq:darcy}
\textcolor{mildred}{- \nabla(K(x) \nabla u(x))} = f(x),
\end{equation}
Commonly used in heat conduction, groundwater flow, and diffusion through materials with spatially varying diffusivity \(K(x)\).

To assess generalization to more complex scenarios, we fine-tune the model on two variants of \cref{eq:darcy}. The first introduces an \textit{reaction} term:
\begin{equation}\label{eq:darcy_absorption}
\textcolor{mildred}{- \nabla(K(x) \nabla u(x))} + \textcolor{mildblue}{q(x) u(x)} = f(x),
\end{equation}
modelling phenomena such as heat loss, chemical reactions, or decay processes.
The second variant adds an \textit{advection} term:
\begin{equation}\label{eq:darcy_reaction}
\textcolor{mildred}{- \nabla(K(x) \nabla u(x))} + \textcolor{mildgreen}{\nabla(v(x) u(x))} = f(x),
\end{equation}
capturing the transport of \(u\) by a velocity field \(v(x)\), relevant in fluid dynamics and pollutant dispersion.

In all cases, the resulting matrices have a condition number on the order of \(10^5\). For the training details, we refer to \cref{sec:bvp_training_details} and for the details regarding the dataset we refer to \cref{sec:data_BVP}.

\section{Datasets for BVP}\label{sec:data_BVP}
For \cref{eq:darcy} the function $K$ is 
$$
K(x) = 1 + \alpha \cos(2 \pi \omega x),
$$
where $\alpha \sim \mathrm{U}[0.25,0.75]$ and $\omega \sim \mathrm{U}[0.01,0.75]$, the function \( K(x)\) models a spatially varying medium with periodic structure.  The parameters \(\alpha\) and \(\omega\) control the amplitude and frequency of variation, allowing tunable complexity. It ensures smoothness and positivity, making it ideal for testing PDE solvers in heterogeneous settings.\\
\begin{figure}[htb]
    \centering
    \includegraphics[width=1\linewidth]{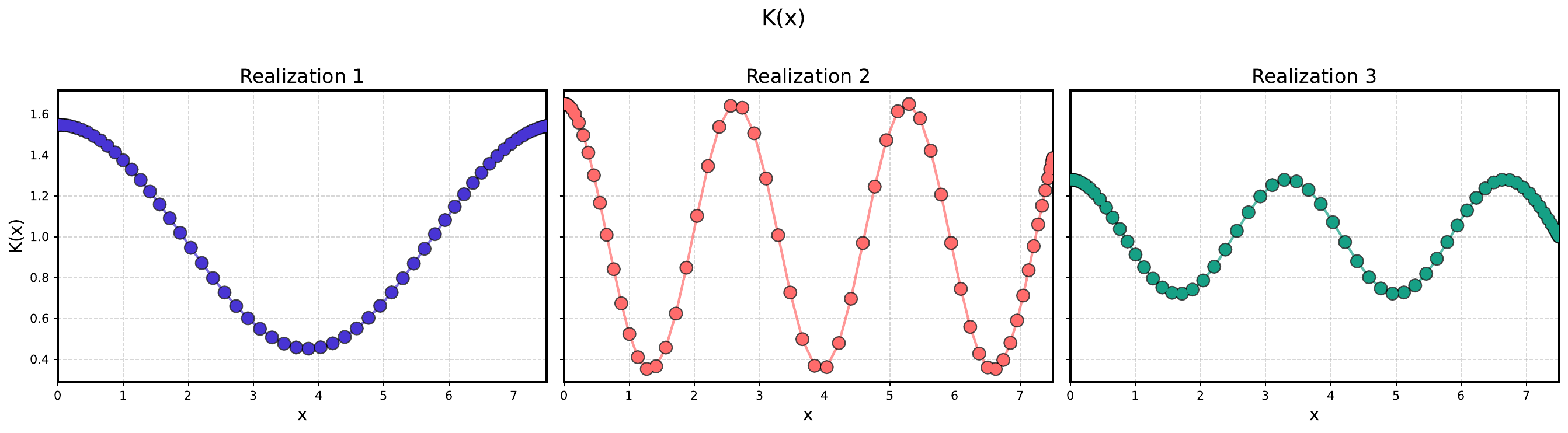}
    \caption{Three different samples of the function $K$}
    \label{fig:bvp_k}
\end{figure}
On the other for, the source term $f(x)$ often represents external forcing, such as heat sources, chemical reactions, or applied loads, distributed across space. The function
\[
f(x) = (1 - \alpha) + \alpha \cdot r(x)
\]
Can be interpreted as a blend between a uniform background source (the constant \( 1 - \alpha \)) and a spatial random fluctuation \( \alpha \cdot r(x) \), where \( r(x) \) is a random field with unit average.
\begin{figure}[htb]
    \centering
    \includegraphics[width=1\linewidth]{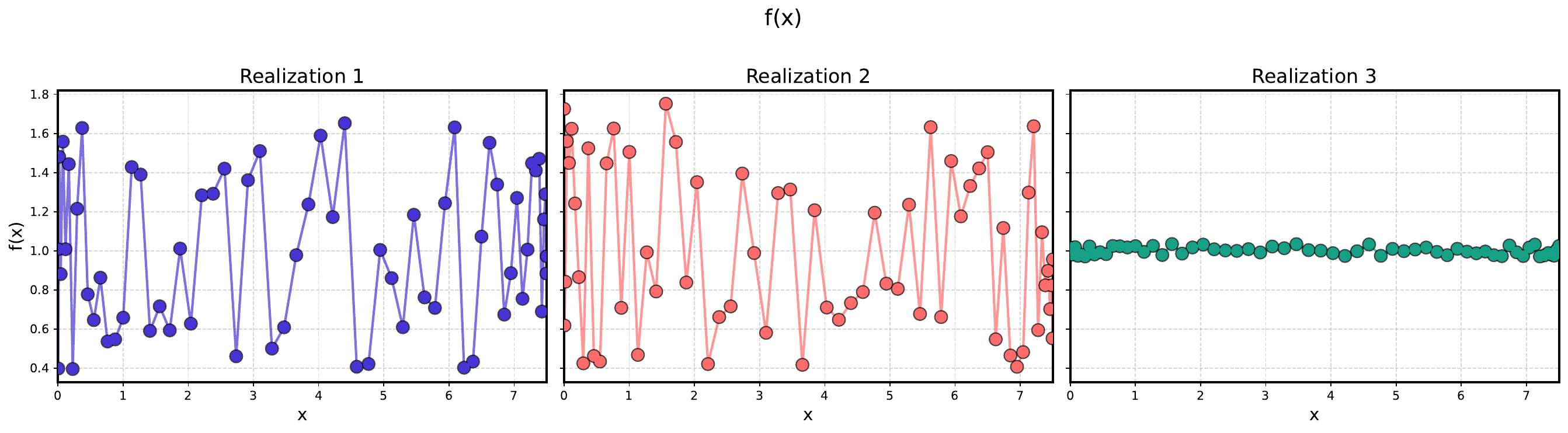}
    \caption{Three different samples of the function $f$}
    \label{fig:bvp_f}
\end{figure}

\begin{figure}[h]
    \centering
    \includegraphics[width=1\linewidth]{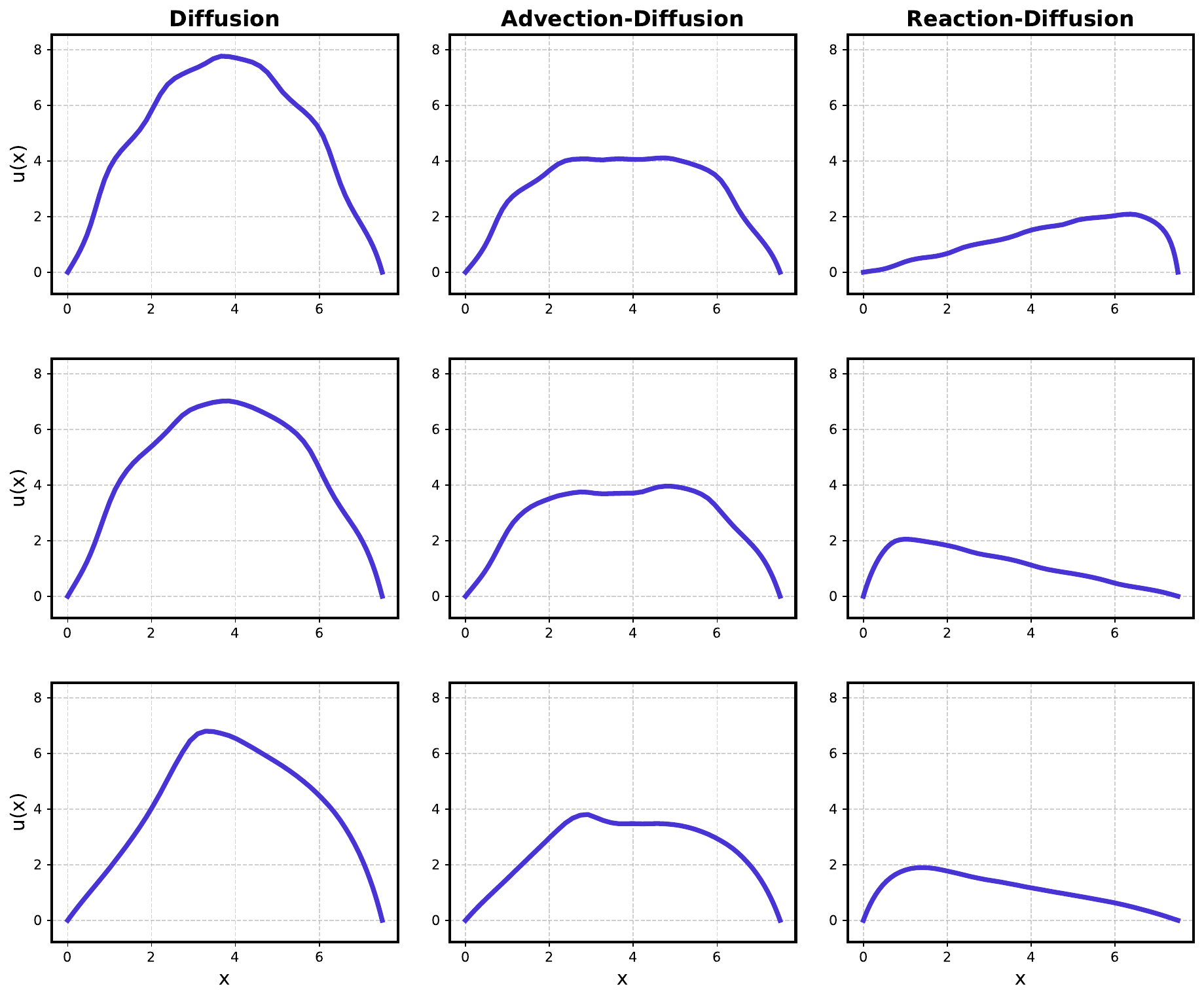}
    \label{fig:bvp_sol}
    \caption{Three different samples for each equation}
\end{figure}

For the \cref{eq:darcy_absorption} we choose the absorption term to be not null only in the interval $[3,4.5]$ where we choose a constant $1/3$, thus as an attractive force that pushes to $0$, the solution $u$. Instead, for the velocity field of \cref{eq:darcy_reaction}, we choose and uniform velocity field $v(x) = \alpha$ where $\alpha \sim U(-2,2)$, based on the sign of the velocity field, the maximum of the solution is moved toward right or left. \\
We plot the three different solutions \cref{fig:bvp_sol}

\section{Training details}\label{sec:bvp_training_details}
For \cref{eq:darcy,eq:darcy_absorption,eq:darcy_reaction}, we aim to predict the approximate solution passing through a basis of functions. The model receives two inputs: for \cref{eq:darcy}, the Chebyshev differentiation matrix based on the spatially variable coefficient \(K\), and for the other two equations, the inputs are dependent on their respective coefficient terms. Furthermore, we concatenate the right-hand side \(f(x)\) into the input matrix, as described in \cref{sec:algebraformer}. The training dataset consists of 50,000 samples, and the test set contains $5,000$ samples; further details on sample construction can be found in \cref{sec:data_BVP}. For the more complex equations \cref{eq:darcy_absorption} and \cref{eq:darcy_reaction}, we train with fewer samples to simulate real-world scenarios where the data for the simpler equations is more abundant. Specifically, we used $250$ training observations for \cref{eq:darcy_absorption} and $1,500$ for \cref{eq:darcy_reaction}, finding that the latter is the minimal number of observations for a good generalisation due to the complexity of the equation. We test the model on 5,000 samples for both equations.

For \cref{eq:darcy}, we train all models for $400$ epochs using the AdamW optimizer with $\beta_1 = 0.9$ and $\beta_2 = 0.95$, applying cosine decay to the learning rate from $10^{-4}$ to $10^{-5}$. For fine-tuning tasks on \cref{eq:darcy_absorption,eq:darcy_reaction}, we use the same learning rate schedule and train for 1,000 epochs when starting from scratch. For fine-tuned models, we use a fixed learning rate of \(5 \times 10^{-5}\) and train for half the epochs. In all experiments, we used a base of 64 elements, resulting in a matrix \(64 \times 64\).

\newpage

\section{Newton Method for $\min_x||Ax-b||_p^p$}\label{sec:newton_theory}

The Newton method is a classical iterative algorithm for solving nonlinear optimization problems~\cite{kelley2003solving}. It is particularly attractive due to its quadratic convergence rate, and it finds widespread application in areas such as interior-point methods for constrained optimization~\cite{waltz2006interior} and large-scale machine learning~\cite{lin2008trust}.

Consider the unconstrained optimization problem:
\[
\min_{x \in \mathbb{R}^n} f(x).
\]
At each iteration, the Newton method computes the next iterate as:
\[
x_{k+1} = x_k - \mathcal{H}_f(x_k)^{-1} \nabla f(x_k),
\]
where \( \nabla f(x_k) \) is the gradient and \( \mathcal{H}_f(x_k) \) is the Hessian matrix evaluated at \( x_k \).

In practice, rather than computing the inverse of the Hessian explicitly, it is common to solve the linear system:
\[
\mathcal{H}_f(x_k) \, p_k = \nabla f(x_k),
\]
and update using \( x_{k+1} = x_k - p_k \). This approach avoids direct inversion but still incurs high computational cost.

Both strategies, explicit matrix inversion and linear system solving, are expensive in terms of both time and resources, especially in high-dimensional settings. This motivates the search for techniques to accelerate the Newton method while preserving its fast convergence properties.

\begin{algorithm}[htb]
\caption{Newton's Method}
\begin{algorithmic}[1]
\Procedure{NewtonMethod}{$f, \nabla f, \mathcal{H}_f, x_0, \text{tol}, \text{max\_iter}$}
    \State $x \gets x_0$
    \For{$k = 0$ to max\_iter}
        \State $g \gets \nabla f(x)$
        \If{$\|g\| < \text{tol}$}
            \State \Return $x$
        \EndIf
        \State $H \gets \mathcal{H}_f(x)$
        \State Solve $H p = g$ for $p$
        \State $x \gets x - p$
    \EndFor
    \State \Return $x$
\EndProcedure
\end{algorithmic}
\end{algorithm}

As discussed, Newton's method is a powerful tool for solving nonlinear optimization problems. We now consider applying it to the following objective:
\[
f(x) = \|Ax - b\|_p^p,
\]
where \( A \in \mathbb{R}^{m \times n} \), \( b \in \mathbb{R}^m \), and \( p \geq 1 \). This function arises in robust regression and data fitting applications, especially when \( p \neq 2 \).

Unlike the \( \ell_2 \)-norm case (where \( p = 2 \)), the \( \ell_p^p \)-norm leads to a nonlinear, non-quadratic objective for general \( p \). As a result, the optimization problem becomes nonlinear and requires iterative methods like Newton's method for efficient minimization.

Let us define the residual vector \( r(x) = Ax - b \). Then the objective can be written as:
\[
f(x) = \sum_{i=1}^m |r_i(x)|^p.
\]

\paragraph{Gradient.} The gradient of \( f(x) \) is given by:
\[
\nabla f(x) = p A^\top \left( |r(x)|^{p-1} \odot \operatorname{sign}(r(x)) \right),
\]
where \( \odot \) denotes elementwise multiplication, and the operations \( |\cdot|^{p-1} \) and \( \operatorname{sign}(\cdot) \) are applied elementwise.

\paragraph{Hessian.} The Hessian \( \mathcal{H}_f(x) \) is:
\[
\mathcal{H}_f(x) = p(p - 1) A^\top \operatorname{diag}\left( |r(x)|^{p - 2} \right) A,
\]
which is valid for \( p > 1 \) and \( r_i(x) \neq 0 \). For values near zero, regularization or smoothing techniques are often needed to ensure numerical stability.

\end{document}